\documentclass{article}
\usepackage{spconf,amsmath,graphicx,xcolor, amssymb, multirow}
\usepackage{hyperref}

\title{User Constrained thumbnail Generation using Adaptive Convolutions}

\name{Perla Sai Raj Kishore\textsuperscript{1}, Ayan Kumar Bhunia\textsuperscript{2}\sthanks{Corresponding Author}, Shuvozit Ghose\textsuperscript{1}, Partha Pratim Roy\textsuperscript{3}}

\address{\textsuperscript{1}Institute of Engineering \& Management, India \hspace{0.1cm} \textsuperscript{2}Nanyang Technological University, Singapore\\ \textsuperscript{3}Indian Institute of Technology Roorkee, India \\
{\tt\small \textsuperscript{2}ayanbhunia@ntu.edu.sg }
}
\begin{document}
%
\maketitle
\begin{abstract}

Thumbnails are widely used all over the world as a preview for digital images. 
In this work we propose a deep neural framework to generate thumbnails of any size and aspect ratio, even for unseen values during training, with high accuracy and precision.
We use Global Context Aggregation (GCA) and a modified Region Proposal Network (RPN) with adaptive convolutions to generate thumbnails in real time.
GCA is used to selectively attend and aggregate the global context information from the entire image while the RPN is used to predict candidate bounding boxes for the thumbnail image.
Adaptive convolution eliminates the problem of generating thumbnails of various aspect ratios by using filter weights dynamically generated from the aspect ratio information.
The experimental results indicate the superior performance of the proposed model\footnote{The source code of the proposed system is publicly available at \url{https://github.com/sairajk/Thumbnail-Generation}} over existing state-of-the-art techniques.

\end{abstract}
\section{Introduction}
\label{sec:intro}

With an increasing number of digital images everyday, it is a need to display the images in any given space efficiently for easy browsing.
This is facilitated by the use of thumbnail images which are a smaller version of the original images that effectively capture and represent the content of the original images.
Thumbnails are widely used in social media platforms and for applications such as image gallery where there is a need to show the images within a fixed resolution display and in an organized manner.

Earlier methods on thumbnail generation \cite{suh2003automatic, ciocca2007self, chen2016automatic} employ a two-stage framework which first predict a salient map specifying the importance of various regions and objects of the input image and then use this for generating a crop for the thumbnail through a region search algorithm.
Although \cite{chen2016automatic} could crop images with a limited set of aspect ratios, it was mentioned it could be infeasible for a given overall saliency threshold value.
Other approaches like \cite{wang2018deep, Li_2018_CVPR} attempt to crop the most aesthetic part of the image.
Huang et al. \cite{huang2015automatic} were the first to address this problem with a direct solution. They used manually selected features and SVM to score a large set of candidate crops. The crop with largest score was finally selected as the thumbnail. But their method was slow and only considered thumbnails of fixed size. 
Esmaeili et al. \cite{esmaeili2017fast}, proposed a fast thumbnail generation algorithm inspired from a object detection framework (F-RCN) \cite{dai2016r} to predict the bounding box coordinates of the thumbnail image. 
They utilize the aspect-ratio information by maintaining a set of filter banks for various aspect-ratio values.

\begin{figure}[t]
\centering
\includegraphics[width=1.0\linewidth]{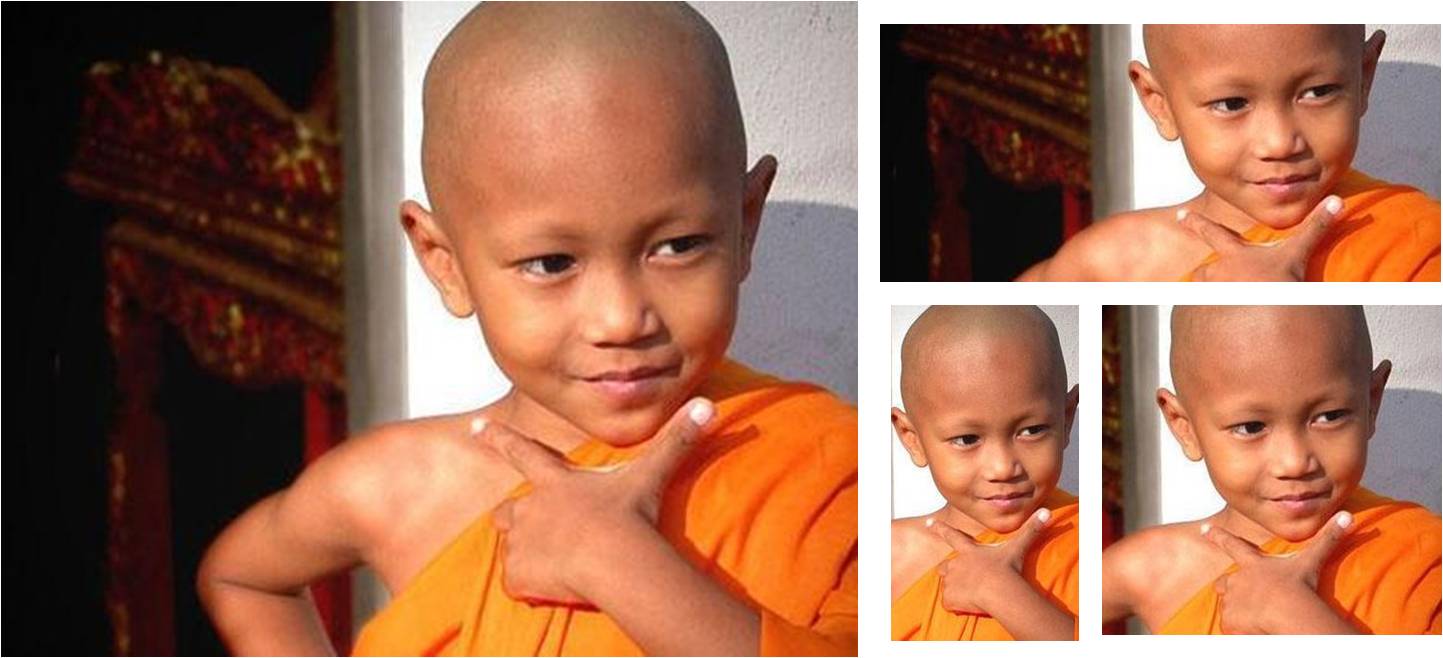}
   \caption{An example of output generated by our model. The original image is shown on the left with the generated thumbnails of different sizes and aspect ratios on the right.}
\label{fig:first}
\end{figure}

\begin{figure*}[t]
\centering
\includegraphics[width=1.0\linewidth]{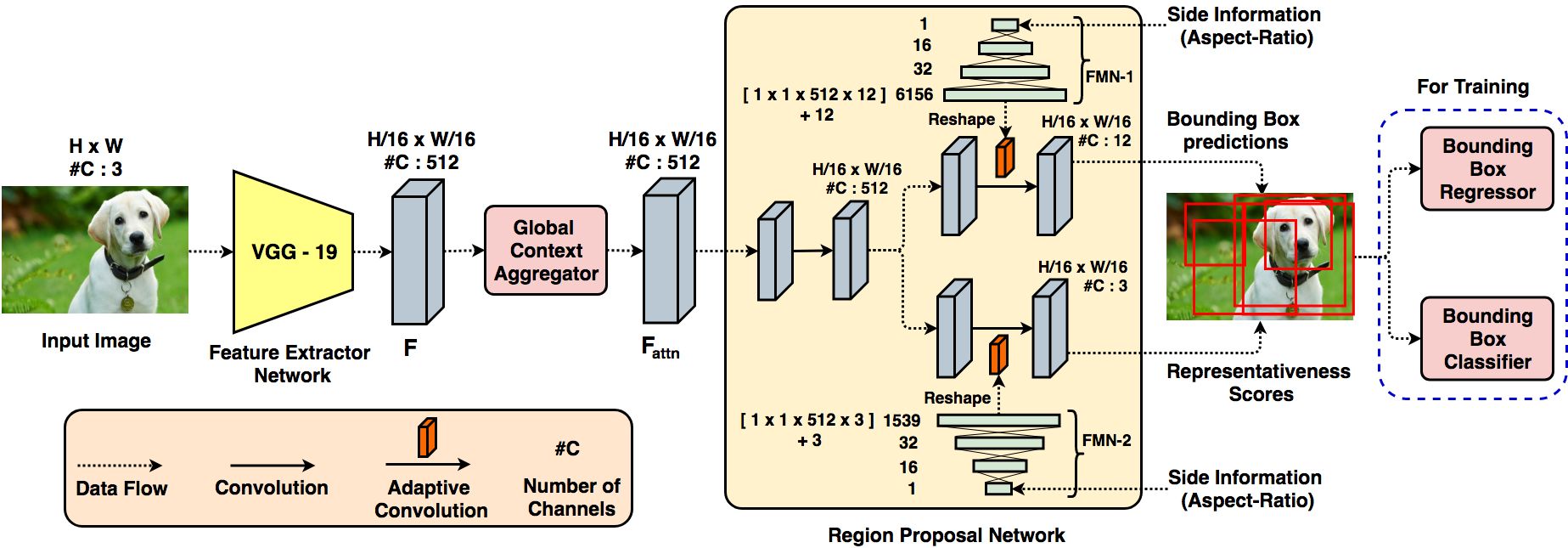}
   \caption{Architecture of the proposed model. The input image is passed through a Feature Extraction Network to generate $F$. The global context is then selectively aggregated at each 
   location to generate $F_{attn}$. A modified RPN is then used to generate candidate bounding boxes and 
   repsentativeness scores. The network is finally trained using Regression and Classification losses.}
\label{fig:architecture}
\end{figure*}

In this work, we propose a deep neural framework which generates a thumbnail of any given size and aspect-ratio for an input image in real time.
We use Global Context Aggregation (GCA) and a Region Proposal Network (RPN) with adaptive convolutions to generate thumbnails of varying sizes with high accuracy and precision.
GCA selectively attends and aggregates the contextual information from the entire image at each context location, increasing the receptive field of the model to the entire image and enhancing the feature representability of the CNNs.
A RPN is then used to generate candidate bounding boxes for the thumbnail image.
The RPN uses adaptive convolutions where the convolution filter weights are generated dynamically according to the aspect ratio of the thumbnail image.
The adaptive convolution layer provides a smooth manifold for the convolution filters to vary and is used to disentangle the variations in shape of the bounding box predictions with respect to aspect ratio.
The smooth manifold for convolution filters also allows us to interpolate and use unseen but similar values of aspect ratio for thumbnails during inference.

In summary, we make the following two contributions :

\noindent
$(1)$ Our method can generate thumbnails of any size and aspect ratio efficiently, even for unseen values during training.

\noindent
$(2)$ Our method is fast and can generate thumbnails in real time. It can process around 12 images per second on a GPU.

\section{Methodolgy}

\subsection{Global Context Aggregation (GCA)}
\label{GCA}

Given a convolutional feature map, $\textbf{F} \in \mathbb{R}^{H \times W \times C}$, with height $H$, width $W$ and number of channels $C$, GCA aims to generate an attention map at each location, over 
$\textbf{F}$ to selectively aggregate the global context information from the entire image.
We employ a recurrent approach, inspired from \cite{liu2018picanet} 
and \cite{visin2015renet}, for this purpose.
A pair of Bidirectional LSTMs is used to scan the given feature map in horizontal and vertical directions from the either ends to aggregate the global context at each spatial location $\langle  h, w \rangle$.
We then apply convolution to generate an intermediate output $Z \in \mathbb{R}^{H \times W \times D}$, where $D = H \times W$. 
Softmax operation is then applied to normalize each of these $D-$dimensional vectors of $Z$ and can be represented as :
\begin{equation}
Z_{norm}^{\langle h, w, i \rangle} = \dfrac{exp(Z^{\langle h, w, i \rangle})}{\sum_{i = 1}^{D} exp(Z^{\langle h, w, i \rangle})}
\end{equation}
where $i$ is the index along the depth of the convolution output $Z$ and $Z_{norm}$ is the normalized feature map.
The features at all positions in $\textbf{F}$ are then weighted summed by $Z_{norm}^{\langle h, w \rangle}$ to generate the attended contextual features $\textbf{F\textsubscript{attn}} \in \mathbb{R}^{H \times W \times C}$: 
\begin{equation}
F_{attn}^{\langle h, w \rangle} = \sum_{i = 1}^{D} Z_{norm}^{\langle h, w, i \rangle f^{i}}
\end{equation}
where $f^{i}$ is the $C-$dimensional feature vector at the $i^{th}$ location (row/column-wise) of $\textbf{F}$.


\subsection{Adaptive Convolutional Layer}

To address the difficulty of generating thumbnails of various aspect-ratios, we make use of adaptive convolutions \cite{kang2016crowd} which model the filter weights as a low-dimensional manifold, parametrized by the side information, within the high-dimensional space of filter weights.
In other words, the filter weights for adaptive convolution are generated dynamically based on some side information (aspect ratio, in this case).
Hence, as the aspect ratio changes, the convolution filter weights also changes accordingly.
This eliminates the need of aspect ratio-specific filter banks as used in \cite{esmaeili2017fast}, making the same convolutional layer adaptive to various values of aspect ratio for the thumbnail image.
This allows us to disentangle aspect ratio-specific outputs and also use unseen but similar values of aspect ratio during inference.

Compared to a normal convolutional layer where the filter weights remain constant after training, the filter weights for the adaptive convolutional layer are dynamically generated by a set of fully connected layers, called the Filter Manifold Network (FMN), with progressively increasing number of neurons and side information as input.
The convolution filter weights are generated by reshaping the output of the FMN into a $4D$ tensor for convolution kernel and a $1D$ vector for bias.
The adaptive convolution layer is more formally defined by, 
$F_{adap} = A(x * g(z; w))$,
where $x$ is the input image or a feature map from previous convolutions, $z$ is the side information, $g(.;w)$ is the filter manifold network with learnable weights $w$ and $A(.)$ is the activation function.

\subsection{Thumbnail Generation}
\label{sec:TG}

We employ a two-step process for generating the thumbnail, a bounding box with user defined aspect ratio is first predicted to crop the region of interest, the cropped image is then rescaled to generate the thumbnail of desired size.

The input image is first passed through a series of convolutional layers, which act as a feature extractor network.
The extracted features, $\textbf{F}$, are then passed through a GCA module
to aggregate the global context information at each location.
These attended contextual features, $\textbf{F\textsubscript{attn}}$, are then passed on to a Region Proposal Network (RPN) to predict candidate bounding boxes for the thumbnail image.
The RPN generates two outputs, bounding box coordinates and the corresponding representativeness scores at each spatial location of the input feature map.
The representativeness score depicts how well the corresponding bounding box represents the original image.
The RPN consists of a convolution layer with $3 \times 3$ kernel and $512$ channels as output, followed by two parallel convolutions using $1 \times 1$ kernel and whose number of output channels depend upon the number of bounding boxes (anchors), $k$, predicted at each location.
One branch of the $1 \times 1$ convolution generates bounding box predictions with $4k$ channels as output ($\langle x_{center}, y_{center}, width, height \rangle$ for each bounding box), whereas the other branch generates the corresponding representativeness scores with $k$ channels as output 
(probability of the corresponding bounding box representing the original image).
We modify the RPN, proposed in \cite{ren2015faster}, to generate the bounding box coordinates and the corresponding representativeness scores constrained by the user defined aspect ratio for the thumbnail image.
For this we use adaptive convolutional layers where the filter weights are dynamically generated from the aspect ratio information.
We make the $1 \times 1$ convolution layers of the RPN adaptive, as these are the layers which are finally responsible for generating the bounding boxes coordinates and representativeness scores for the target thumbnail image.

We use the attended contextual features, $\textbf{F\textsubscript{attn}}$, instead of directly using the features, $\textbf{F}$, from the feature extractor network for bounding box predictions.
This is vital for a task like thumbnail generation where the entire global context is important for generating the thumbnail, as opposed to object detection tasks where only the local neighborhood context is important.
GCA allows the RPN to \emph{look} at the entire image in a selective manner before making the bounding box predictions at each location and to assess how well the predicted bounding boxes represent the original image (through representativeness scores).
GCA also allows us to use a lighter feature extractor network (in contrast to ResNet-101 \cite{He2016DeepRL} used by \cite{esmaeili2017fast}) by increasing the receptive field, filtering information (by attention) and enhancing the feature representabity of the CNNs.
As a result, GCA plays a crucial role in reducing the memory footprint while improving the overall accuracy and speed of the network.

We predict a total of 3 bounding boxes $($i.e. $k=3)$, with same aspect ratio but different scales (with box areas of $128^2$, $256^2$ and $512^2$ pixels), at each context location of 
$\textbf{F\textsubscript{attn}}$.
The aspect ratio is determined by the convolution kernel generated dynamically from the user input.
We employ a similar approach as used in \cite{ren2015faster} for training the network.
We use regression loss for bounding box predictions and classification loss for the probability scores.
The candidate bounding boxes are first classified as `positive' or `negative' depending upon their Intersection-over-Union (IoU) scores with the ground truth bounding box according to the thresholds described in \cite{ren2015faster}.
We then randomly sample 256 candidate bounding boxes to form a balanced mini-batch of positive and negative boxes.
It is used to calculate the classification loss $(L_{cls})$ using binary cross entropy.
We then only use the positive bounding boxes to calculate the regression loss $(L_{reg})$. For $L_{reg}$, we use smooth $L_1$ loss defined in \cite{girshick2015fast}.
The overall loss function is given by,
\begin{equation} \label{eq:lrpn}
L(p, b) = \dfrac{1}{N_{cls}} \sum_{i} L_{cls}(p_{i}, \widehat{p_{i}}) + \lambda \dfrac{1}{N_{reg}} \sum_{i} \widehat{p_{i}} L_{reg}(b_{i}, \widehat{b_{i}})
\end{equation}
where $i$ is the index of the bounding boxes in the mini-batch and $p_{i}$ is the predicted probability of bounding box being representative of the original image.
The ground-truth label $\widehat{p_{i}}$ is 1 if the bounding box is positive, and is 0 if the bounding is negative.
The predicted bounding box coordinates are represented by $b_{i}$, and that of the associated ground truth box by $\widehat{b_{i}}$.
The two losses are normalized by $N_{cls}$ and $N_{reg}$ and are weighted by $\lambda$.
During inference, the bounding box with highest representativeness score is selected and rescaled to generate the target thumbnail.

\begin{figure*}[t]
\centering
\includegraphics[width=0.919\linewidth]{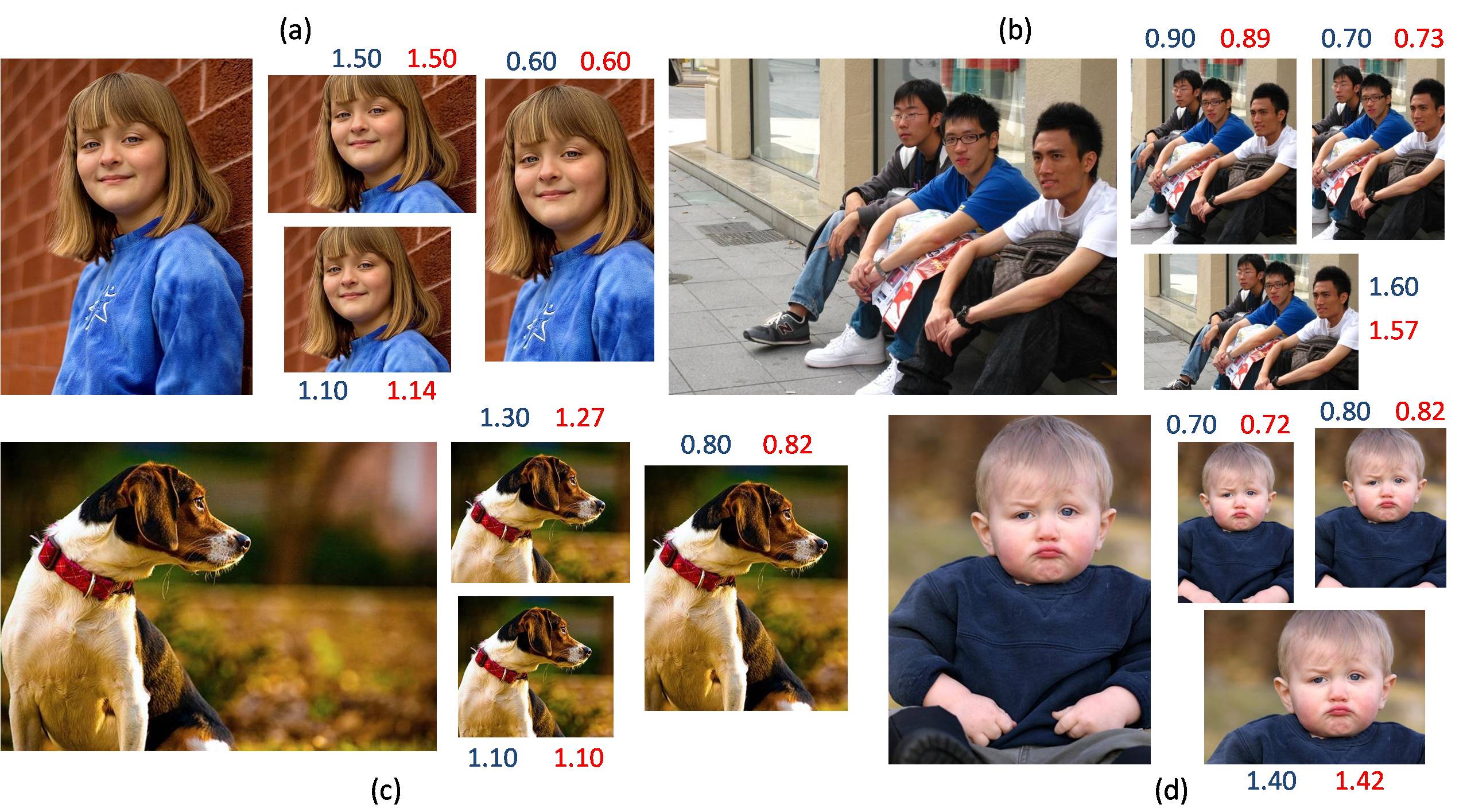}
   \caption{Thumbnails generated by our model on the test set. The original image is shown on the left with the generated thumbnails on the right. The query aspect ratio is given in blue and the aspect ratio of the generated thumbnail is given in red.}
\label{fig:comparision}
\end{figure*}

\section{Experiments}

\textbf{Implementation and Training details:}
For our experiments we use the dataset released in \cite{esmaeili2017fast}.
The dataset contains total of 70,048 thumbnail annotations over 28,064 images taken from the photo-quality dataset \cite{luo2011content}. 
The dataset is split into 63043 annotations (from 24,154 images) for training and 7,005 annotations (from 3,910 images) for testing.
All images are resized to have a maximum height of 650 and a maximum width of 800 pixels.
The size of the annotated thumbnails vary from 32 to 200 pixels and the aspect ratio varies from 0.5 to 2.

The entire network is trained in an end-to-end manner supervised by the loss function stated in Equation \ref{eq:lrpn} with $\lambda = 10$.
We use VGG-19 \cite{simonyan2014very} as the feature extractor network with weights initialized by pretraining the model for ImageNet classification \cite{russakovsky2015imagenet}.
The features, $\textbf{F}$, are extracted from the \texttt{conv5\_4} layer of the VGG-19 network and all the layers upto \texttt{conv5\_4} are fine-tuned during training.
All other weights are initialized with values sampled from zero-mean Gaussian distribution with standard deviation 0.02.
We use Adam Optimizer \cite{DBLP:journals/corr/KingmaB14} for training with learning rate and momentum set to 0.001 and 0.9 respectively.
The model is implemented using the TensorFlow library and all experiments are conducted on a single NVIDIA Tesla K80 GPU. 

\textbf{Baselines and Evaluation Metrics:}
Our model is compared against the following 2 state-of-the-art image cropping methods: Fast-AT \cite{esmaeili2017fast} and Aesthetic Image Cropping (AIC) \cite{wang2018deep}.
Fast-AT maintains a set of aspect ratio-specific convolutional filters 
to generate thumbnails of different aspect ratios.
Whereas, \cite{wang2018deep} crops the most aesthetic part of the image irrespective of the target image aspect ratio.
We use the following evaluation metrics defined in \cite{huang2015automatic} and \cite{esmaeili2017fast} to compare our model to other methods: Center offset (CO), Rescaling factor (RF), Intersection over Union (IoU), Aspect ratio mismatch (ARM), Hit Ratio $(h_r)$ and Background Ratio $(b_r)$.

\textbf{Performance Analysis:}
A quantitative comparison of the proposed model to the other methods for various evaluation metrics is shown in Table \ref{tab:performance} and some qualitative results generated by our model are shown is Figure \ref{fig:first} and Figure \ref{fig:comparision}.
It is evident from the results in Table \ref{tab:performance} that our model achieves the best results for all evaluation metrics and outperforms the existing state-of-the-art methods for thumbnail generation.
We observe a significant drop in ARM score which is attributed to adaptive convolutions, which help us generate bounding boxes with the given aspect ratio with high precision.
Fast-AT \cite{esmaeili2017fast} works moderately well due discretized aspect ratio-specific filters which approximate the target aspect ratio.
Aethetic Image Cropping \cite{wang2018deep} performs poorly as it crops the most aesthetic region irrespective of the aspect ratio of the target image.

\section{Conclusion}

\begin{table}[]
\centering
\resizebox{8.6cm}{!}{
\begin{tabular}{|c||c|c|c|c|c|c|}
\hline
\textbf{Model} & \textbf{CO} & \textbf{RF} & \textbf{IoU} & \textbf{ARM} & $h_r$ & $b_r$ \\ \hline \hline
AIC \cite{wang2018deep} & 89.4 & 1.402 & 0.54 & 0.213 & 65.6\% & 43.1\% \\ \hline 
Fast-AT \cite{esmaeili2017fast} & 55.0 & 1.148 & 0.68 & 0.010 &  83.7\% & 37.1\% \\ \hline
\textbf{Ours} & 51.8 & 1.032 & 0.77 & 0.0013 & 87.6\% & 30.3\% \\ \hline
\end{tabular}
}
\caption{Evaluation Metrics evaluated on different thumbnail generation methods.}
\label{tab:performance}
\end{table}

In this work, we proposed a deep neural framework to generate thumbnails of variable aspect ratio, even for unseen values during training, with high accuracy and precision.
We used Global Context Aggregation and a modified Region Proposal Network with adaptive convolutions to generate state-of-the-art results for automatic thumbnail generation.
The proposed model requires significantly lesser number of parameters and can generate thumbnails in real time.
We finally demonstrated the superior performance of our model over existing methods by qualitative and quantitative results.

\bibliographystyle{IEEEbib}
\bibliography{strings,refs}

\end{document}